\documentclass{article}

\usepackage{arxiv}

\usepackage[utf8]{inputenc} 
\usepackage[T1]{fontenc}    
\usepackage{hyperref}       
\usepackage{url}            
\usepackage{booktabs}       
\usepackage{amsfonts}       
\usepackage{nicefrac}       
\usepackage{microtype}      
\usepackage{lipsum}
\usepackage{graphicx}
\graphicspath{ {./images/} }
\graphicspath{{./Figures/}}

\usepackage{doi}
\usepackage{booktabs}
\usepackage{enumitem,kantlipsum}
\usepackage{amssymb}
\usepackage{bm}

\usepackage[accsupp]{axessibility}  
\usepackage{cleveref}

\title{Success Probability in Multi-View Imaging}

\author{
 Vadim Holodovsky \\
  Viterbi Faculty of Electrical Engineering,\\
  Technion - Israel Institute of Technology,\\
  Haifa, Israel \\
  \texttt{vholod@technion.ac.il} \\
   \And
 Masada Tzabari \\
  Viterbi Faculty of Electrical Engineering,\\
  Technion - Israel Institute of Technology,\\
  Haifa, Israel \\
  \texttt{masada.tzemach@mail.huji.ac.il} \\
  \And
 Yoav Schechner \\
  Viterbi Faculty of Electrical Engineering,\\
  Technion - Israel Institute of Technology,\\
  Haifa, Israel \\
  \texttt{yoav@ee.technion.ac.il} \\
  \And
 Alex Frid \\
  Norman \& Helen Asher Space Research Institute,\\
  Technion - Israel Institute of Technology,\\
  Haifa, Israel \\
  \texttt{alex.frid@gmail.com} \\
 \And
 Klaus Schilling \\
  Zentrum für Telematik e.V., \\
  Würzburg, Germany \\
  \texttt{klaus.schilling@telematik-zentrum.de} 
}

\begin{document}
\maketitle
\begin{abstract}
Platforms such as robots, security cameras, drones and satellites are used in multi-view imaging for three-dimensional (3D) recovery by stereoscopy or tomography. Each camera in the setup has a field of view (FOV). Multi-view analysis requires overlap of the FOVs of all cameras, or a significant subset of them. However, the success of such methods is not guaranteed, because the FOVs may not sufficiently overlap. The reason is that pointing of a camera from a mount or platform has some randomness (noise), due to imprecise platform control, typical to mechanical systems, and particularly moving systems such as satellites. So, success is probabilistic. This paper creates a framework to analyze this aspect. This is critical for setting limitations on the capabilities of imaging systems, such as resolution (pixel footprint), FOV, the size of domains that can be captured, and efficiency. The framework uses the fact that imprecise pointing can be mitigated by self-calibration - provided that there is sufficient overlap between pairs of views and sufficient visual similarity of views. We show an example considering the design of a formation of nanosatellites that seek 3D reconstruction of clouds. 
\end{abstract}


\section{Introduction}
\label{sec:intro}

\begin{figure}[t]
  \centering
  \includegraphics[height=3.5cm]{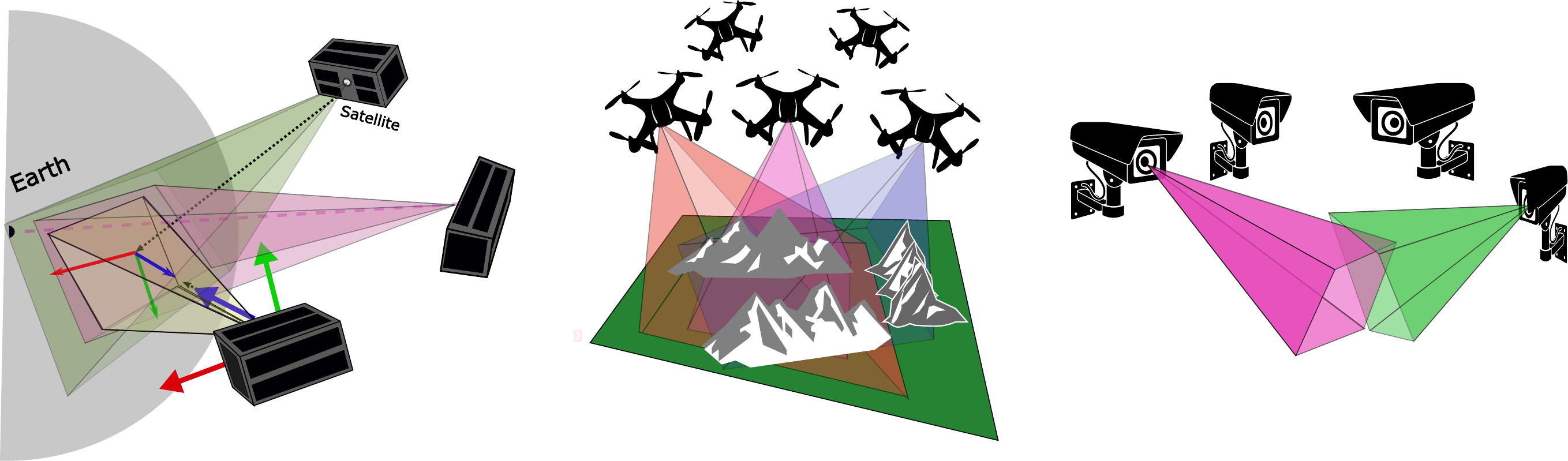}
  \caption{Pointing errors across multi-view platforms: [Left] Satellites having alignment challenges during Earth observation. [Middle] Drones having orientation noise. [Right] The same phenomenon can be seen in ground-based (e.g, security) cameras.
  }
  \label{fig:scenes}
\end{figure}

Multi-viewing imaging systems are common (Fig.~\ref{fig:scenes}) and include: Rigs of security or other monitoring cameras observing a volumetric domain; Cooperating drones observing the same scene, which may be volumetric or a surface domain;
or a fleet of satellites in a formation observing a thin atmospheric domain just above the Earth's surface. Multi-view setups also include a moving platform observing sequentially any such domain. 
Multi-view setups enable three-dimensional (3D) surface reconstruction and computed tomography (CT) outdoors, including CT of turbulence~\cite{Tomography2014,shaul2022tomography} or clouds~\cite{Levis_2017_CVPR,levis2020multi,ronen2022variable,forster2021toward}, and statistical tomography of microscopic organisms~\cite{levis2018statistical}.
Deep-neural networks (DNNs) currently advance large-scale CT~\cite{levis2022gravitationally,bruning2024artificial,lin2023volumetric,ronen2022variable}. DNNs also advance surface recovery based on multi-view stereo~\cite{gu2020cascade,laga2020survey,liu2023deep,kong2024robodepth,voynov2023multi, isprs-archives-XLVIII-1-W2-2023-1021-2023}.

As cameras and agile platforms become very affordable (not least by cameras being in smartphones), such uses are set to broaden, particularly using compact, low-resource platforms. 
However, the success of 3D recovery relies, first of all, on having the right data, that is, having the domain of interest in overlapping views from a sufficient number of viewpoints.  
This is not guaranteed. Ideally, a platform carrying a camera would point in a direction it is commanded to, with perfect accuracy. But platforms used to position and orient cameras invariably introduce pose knowledge errors. Each platform has limited accuracy in pointing and pose sensing. Consequently, there is a random error between the actual pointing direction and a commanded direction~\cite{steyn2020stability} given to the platform aiming system. The standard deviation of this error is termed {\em absolute pointing error} (APE).  This problem is exacerbated in small, low-resource platforms. The cause is mechanical inaccuracies and feedback, especially in moving platforms such as small satellites and drones, which rely
on an
attitude determination and control system (ADCS)~\cite{steyn2020stability}.  
Therefore, the success of multi-view imaging is probabilistic. 

This paper creates a framework to analyze the success probability. This is critical for setting limitations on the capabilities of imaging systems, such as resolution (pixel footprint), FOV, the size of domains that can be captured, and efficiency. To the best of our knowledge, this kind of analysis has not been done before.

The framework relates to self-calibration. Geometric self-calibration is thoroughly developed in computer vision~\cite{hartley2003multiple,zhang2019leveraging, jovanovic1999level}. It can help detect pose errors, e.g, using structure from motion~\cite{zhang2019leveraging,agarwal2011building,delmerico2018benchmark,bryson2008observability} {\em after image data had already been acquired}~\cite{gomez2022experimental,jovanovic1999level,zhang2019leveraging,hu2009geometric,cui2015efficient}. However, extrinsic geometric self-calibration requires the following: 
\begin{enumerate}[label=(\alph*)]
    \item 
    A sufficient overlap of the fields of view (FOVs), specifically between FOVs corresponding to nearest-neighboring viewpoints.
    \item 
    A sufficient number of geometric features in the region of the overlapping FOVs. Each camera should view identifiable features, that can be identified from at least one other view. 
\end{enumerate}  
The framework of this paper relies on these principles. Together with a noise model of pointing, we derive the probability for self-calibration to succeed, and consequently, for 3D reconstruction to succeed. 

As a case study, we apply this analysis to considerations that should be addressed by designers of space missions. Due to their relatively low production and launch costs, there is a growing trend towards deploying formations of small satellites for Earth observation missions~\cite{schilling2019cloudct,choi2022geoscan,kleinschrodtatom,dauner2023visual,schilling2021small,scharnagl2021netsat}.
However, they often trade-off affordability and scalability for limited pointing capabilities, when compared to large satellites~\cite{diner1998multi,genkova2007cloud,Breon2005,platnick2016modis,parol2004review,li2018directional}. This presents unique challenges for 3D imaging tasks, where precise alignment and overlap of images from multiple spaceborne views are crucial for accurate data analysis.


\section{Overlap Measures}
\label{sec:overlap_sim}

\begin{figure}[t]
  \centering
  \includegraphics[height=5.8cm]{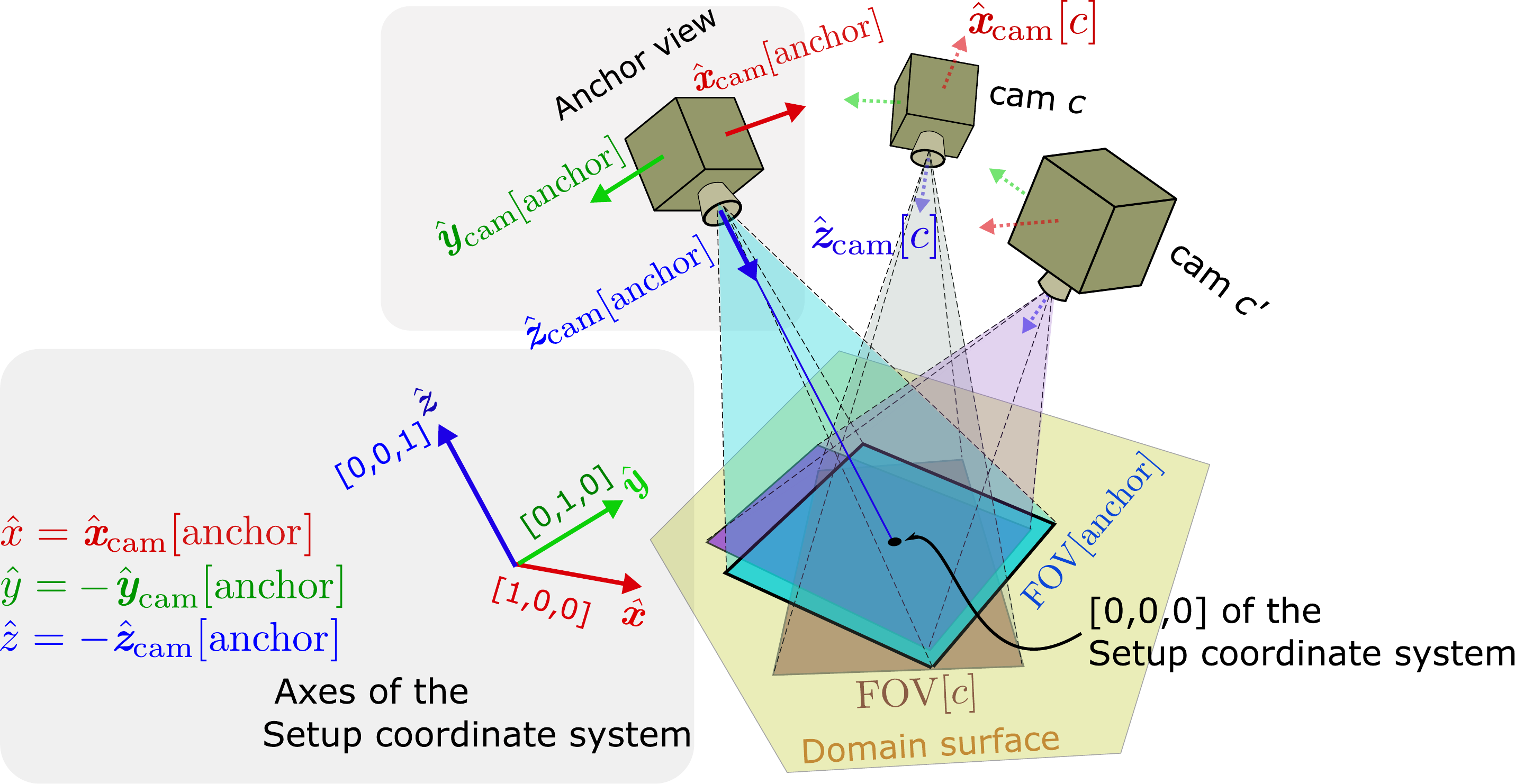}
  \caption{A general multi-view setup. One of the cameras is the anchor. Its axes and location correspond to the setup coordinate system. Each camera has a FOV on the domain surface.
  }
  \label{fig:general_setup}
\end{figure}

Fig.~\ref{fig:general_setup} illustrates the concept of overlap between FOVs on a reference surface. Let ${c}$ denote an index of a camera. 
Assume identical cameras within the setup. Each camera has a frustum of vision: a pyramid-shaped volume that starts at the camera optical center and extends in 3D, as in Fig.~\ref{fig:general_setup} and \cite{szeliski2022computer}. The intersection of the frustum of camera $c$ with a reference surface creates a polygonal ${\rm FOV}[c]$, which we describe in \cref{sec:fov}.
Let $N_{\rm cam}$ be the number of cameras in a multi-view setup. The setup yields a set of polygons $\Psi = \big\{{\rm FOV}[c]  \big\}_{c=1}^{N_{\rm cam}}$. Set one camera as an {\em anchor}. Often, users are most interested in observing the finest details of an object of interest. Therefore, the anchor is the camera closest to the domain of interest. Consequently, $|{\rm FOV}[{\rm anchor}]|\le |{\rm FOV}[c]|,\;\forall c$. Overlap-related terms are defined as follows:
\begin{itemize}
        \item
    \textit{Absolute-Overlap} is the total area of overlap of FOVs of $N_{\rm cam}$ views:
    \begin{equation}
    \label{eq:ao}
        {\rm AO} = \left |\bigcap_{c=1}^{N_{\rm cam}} {\rm FOV}[c]\right |.
    \end{equation}

    \item 
    \textit{Relative-Overlap} of $N_{\rm cam}$ views:
    \begin{equation}
    \label{eq:ro}
        {\rm RO} = \frac{\rm{AO}}{|{\rm FOV}[{\rm anchor}]|} \leq 1.
    \end{equation}

    \item 
    \textit{Relative pair-wise overlap} of views ${c}$ and ${c'}$ is
    \begin{equation}
    \label{eq:ro_2views}
        {\rm RO}_{{c}, {c'}}= \frac{ |{\rm FOV}[c] \cap {\rm FOV}[c']|}{{\rm min} \{ |{\rm FOV}[c]|, |{\rm FOV}[c']| \}} \leq 1.
    \end{equation}
    
    \item
    \textit{Mean Absolute-Overlap} $\overline{\rm AO} = \langle  \rm AO \rangle$, is the average of ${\rm AO}$, resulting from $N_{\rm MC}$ random samples of $\Psi$.
     In \cref{sect:formation}, we detail how to obtain statistics of these terms by using Monte Carlo sampling.

    \item 
    \textit{Mean Relative-Overlap} $\overline{\rm RO} = \langle  \rm RO \rangle $, is the RO averaged over $N_{\rm MC}$ random samples of $\Psi$, each generated by Monte-Carlo.

\end{itemize}  

\cref{sec:intro} lists conditions for successful self-calibration of geometric parameters. Here, we elaborate on how to measure the probability of meeting issue (a). A key term is {\em connected} view pairs. Let $\mu_{c, c'}$ be the angular difference between cameras indexed $c$ and $c'$. The visual similarity between views $c$ and $c'$ depends on $\mu_{c, c'}$.
For successful geometric self-calibration of an image pair, we need a sufficient number of visually similar image features. Therefore, it is important that the following coexists:
\begin{itemize}
\item A small $\mu_{c, c'}$, in order to have sufficient similarity. This means that visual features are similar between views $c$ and $c'$.
\item A large ${\rm RO}_{c, c'}$, in order to have a sufficient number of matching visual features within the domain.
\end{itemize}
If $\mu_{c, c'}$ is large, images $c$ and $c'$ may significantly differ, resulting in very few or null matching feature points between the views. 
We assume that for a threshold $\mu_{\rm max}$, if $\mu_{c, c'}\le\mu_{\rm max}$, then features are likely to be matched.

Let ${\rm T}\in[0,1]$ be a unitless threshold; representing the minimally required ${\rm RO}$ of any pair of views ${c}$ and ${c'}$. In other words, we need ${\rm RO}_{c, c'}\ge{\rm T}$ for successful pair-wise self-calibration. 
We distinguish between the above critical conditions. 
The similarity issue depends on the {\em scene}, not the setup. For example, colorful bright particles that shine isotropically are visually identified even from mutually very distant views. In contrast, features on complex 3D shapes such as trees and turbulent clouds are hard to identify and match visually, even from angularly close views. 
The aspect ${\rm RO}_{c, c'}\ge{\rm T}$ is a property of the setup. This issue can be engineered by the optics.

\subsection{Connected graph}
\label{sec:con_graph} 
\begin{figure}[h]
  \centering
  \includegraphics[height=5.3cm]{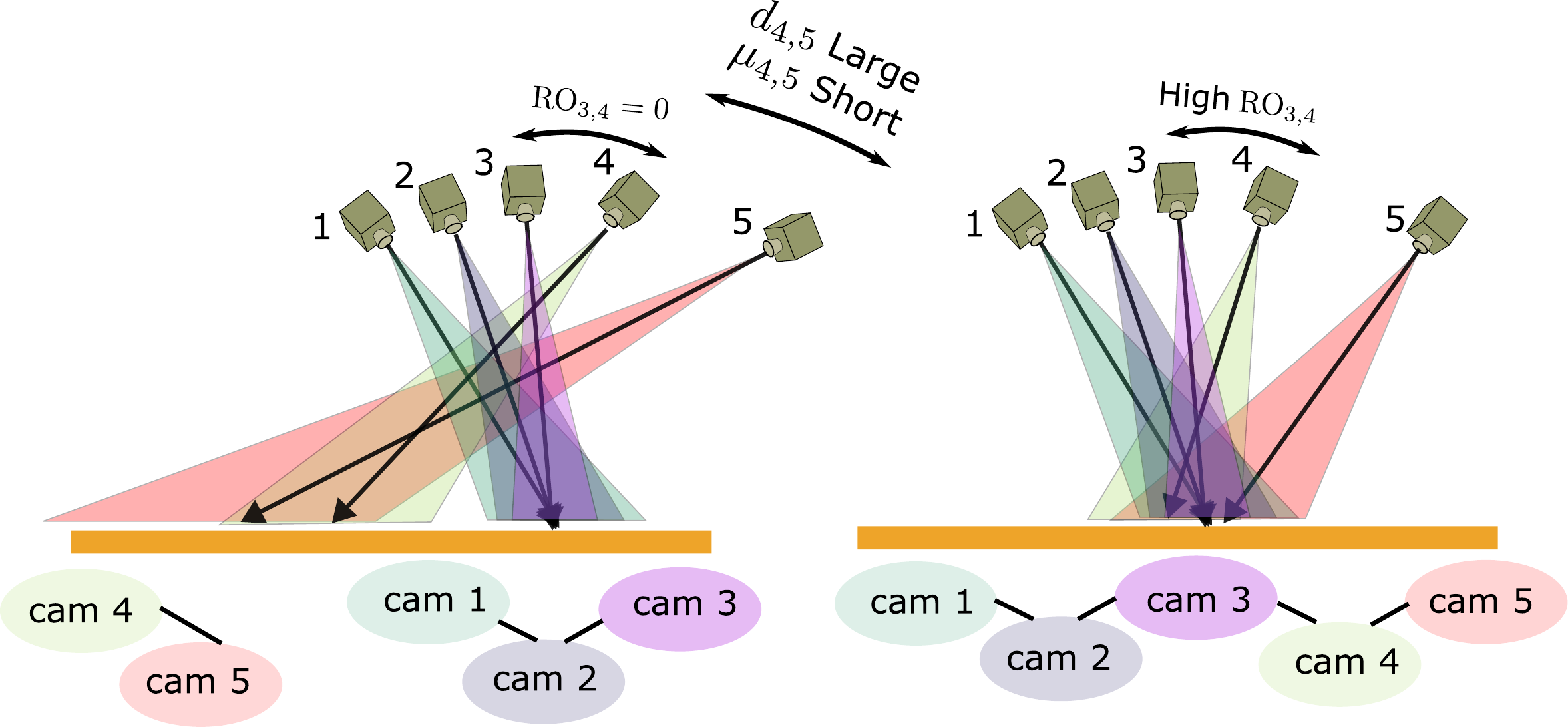}
  \caption{ [Left] Disconnected views (nodes). In this example, there are two graph components, ${\rm B}({\rm Q}=3)$ and ${\rm B}({\rm Q}=2)$.
    [Right] A graph of a connected set of views. Connection is marked by an edge. An edge between nodes $c$ and $c'$ means that ${\rm RO}_{c, c'}>{\rm T}$ and $\mu_{c, c'} \le \mu_{\rm max}$. The graph has a single component ${\rm B}({\rm Q}=5)$.
  }
  \label{fig:graph}
\end{figure}

Let ${\rm Q}$ be the number of viewpoints to undergo geometric self-calibration. To achieve geometric self-calibration for each of the ${\rm Q}$ views, any view must have a significant FOV overlap with some other view, i.e., the setup must have pair-wise overlaps. 
Here we relate connected pair-wise overlaps to a {\em connected graph} concept.
Let ${\rm B}({\rm Q})$ denote a {\em connected} set of ${\rm Q}$ views within the setup, i.e., ${\rm B}({\rm Q})$ is a cluster of ${\rm Q}$ views. 
Fig.~\ref{fig:graph} shows the concept of ${\rm B}({\rm Q})$ as a connected set represented by a graph. Graph edges between two views (graph nodes) $c$ and $c'$ express that the requirements on the two critical conditions above are met. 
Fig.~\ref{fig:graph}[Right], shows a connected graph having ${\rm B}({\rm Q}=5)$. There, pair-wise self-calibration propagates across all views. 
Fig.~\ref{fig:graph}[Left] shows a graph of two disconnected components. In the ${\rm B}({\rm Q}=3)$ cluster, geometric self-calibration can use three views of the setup. The other two views cannot be self-calibrated with the ${\rm B}({\rm Q}=3)$ cluster.

\begin{figure}[h]
  \centering
  \includegraphics[height=3.5cm]{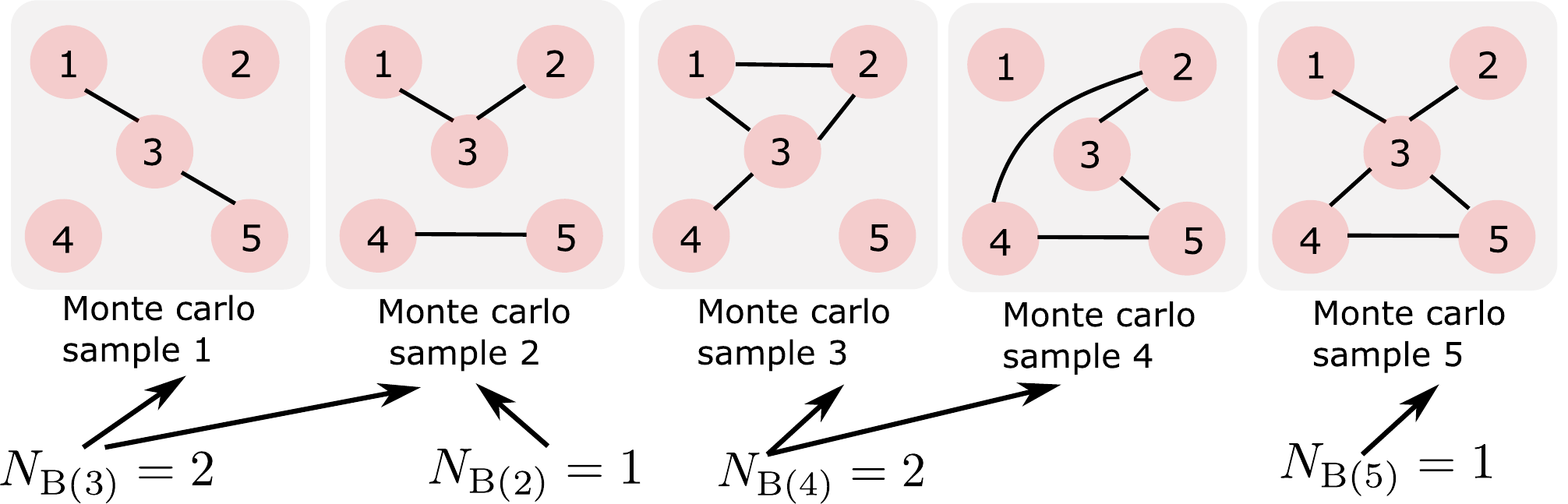}
  \caption{Example of estimation of components of ${\rm P}_{\rm calib}$ via Monte Carlo with $N_{\rm MC}=5$.
  }
  \label{fig:graph_tutorial}
\end{figure}

Define ${\rm P}_{\rm calib} \big( {\rm T} \mid |{\rm FOV}[{\rm anchor}]|, {\rm APE} , {\rm Q} \big)$ as the probability that a set of {\em at least} ${\rm Q}$ views can be self-calibrated. To estimate ${\rm P}_{\rm calib} \big( {\rm T} \mid |{\rm FOV}[{\rm anchor}]|, {\rm APE} , ${\rm Q}$\big)$, we use Monte-Carlo, as we describe in \cref{sect:resutls}.
Let $N_{{\rm B}({\rm Q})}$ be the number of Monte Carlo samples, in which any ${\rm Q}$ views create a cluster ${\rm B}({\rm Q})$. \cref{fig:graph_tutorial} shows an example using $N_{\rm MC}=5$. Here, there is one sample with ${\rm B}(3)$, one sample with ${\rm B}(5)$, two samples with ${\rm B}(3)$, and two samples with ${\rm B}(4)$. The probability to have component of ${\rm B}(5)$ is $\frac{{\rm B}({\rm Q})}{N_{\rm MC}}=1/5$, and component of ${\rm B}(4)$ is 2/5. The probability of getting a cluster with {\em at least} three nodes is ${\rm B}(3)$+${\rm B}(4)$+${\rm B}(5)$ = 1.
To generalize, the probability to have a ${\rm B}({\rm Q})$  component is $N_{{\rm B}({\rm Q})}/N_{\rm MC}$. Therefore,
\begin{equation}
\label{eq:p_calib}
    {\rm P}_{\rm calib} \big( {\rm T} \mid |{\rm FOV}[{\rm anchor}]|, {\rm APE} , {\rm Q} \big) \approx \sum_{{\rm Q'} = {\rm Q}}^{N_{\rm cam}}\frac{N_{{\rm B}({\rm Q'})}}{N_{\rm MC}}.
\end{equation}


Any Monte Carlo sample creates a random state $\Psi$.
Using ${N_{\rm MC}}$ samples, we can estimate ${\rm P}_{\rm calib}$. \cref{eq:p_calib} expresses the probability of achieving geometric self-calibration using ${\rm Q}$ views.

\section{Pointing noise}
\label{sec:point_noise}

\subsection{Coordinate systems}
\label{sec:coord_systems}
Two coordinate systems (Fig.~\ref{fig:general_setup}) are referred to throughout the paper. 
\begin{enumerate}
\item The {\em setup coordinate} system has an origin on
the domain surface, typically the ground. It is right-handed. The axes are labeled $\bm{\hat{x}}$, $\bm{\hat{y}}$, and $\bm{\hat{z}}$. 

\item The {\em camera coordinate system} is right-handed.
Per camera $c$, the axes of the camera coordinate system are denoted ${\hat{\bm x}}_{\rm cam}[c],\,{\hat{\bm y}}_{\rm cam}[c],\,{\hat{\bm z}}_{\rm cam}[c]$. 
Axis ${\hat{\bm z}}_{\rm cam}[c]$ aligns with the optical axis of the camera $c$. Axes ${\hat{\bm x}}_{\rm cam}[c]$
and ${\hat{\bm y}}_{\rm cam}[c]$ are parallel to the pixel rows and columns of the sensor of camera $c$, respectively~\cite{tzabari2022advances}.
 
\end{enumerate}

The anchor camera relates these two coordinate systems. In our context, the origin of the setup coordinate system is the intersection of a line of sight from the anchor camera (in the direction of ${\hat{\bm z}}_{\rm cam}[\rm anchor]$) with the reference surface (Fig.~\ref{fig:general_setup}). Therefore, $\bm{\hat{z}}=-{\hat{\bm z}}_{\rm cam}[\rm anchor]$, $\bm{\hat{x}}={\hat{\bm x}}_{\rm cam}[\rm anchor]$, and $\bm{\hat{y}}=-{\hat{\bm y}}_{\rm cam}[\rm anchor]$.

Let ${\bf R}[c]$ denote a camera $c$ rotation matrix from the camera coordinate system to the setup coordinate system.
Let us denote a general vector in the setup coordinate system by ${\bm \upsilon}$ and a general vector in the camera coordinate system of camera $c$ by ${\bm \upsilon}^{\rm cam}[c]$, then
\begin{equation}
    {\bm \upsilon} = {\bf R}[c] \; {\bm \upsilon}^{\rm cam}[c].
\label{eq:omega_exam}
\end{equation}

\subsection{Ideal pointing}
\label{sec:ideal_point}

Let ${\bm u}_{\text{cam}}[c] = \big[ X_{\rm cam}[c], Y_{\rm cam}[c], Z_{\rm cam}[c] \big]^{\top}$ denote the location of camera $c$ in the setup coordinate system, where $()^{\top}$ denotes transposition.
Suppose all cameras rotate to point to the origin of the setup coordinate system [0,0,0][km] (Fig.~\ref{fig:general_setup}). To achieve this, define matrix ${\bf R}[c]$ using the {\em LookAt transform}~\cite{kessenich2016opengl}, as follows. 
First, set all the cameras such that the 3D vector along the sensor pixel columns is $\hat{\bm y}_{\rm cam}[c]=[0,-1,0]^{\top}$, $\forall c$. Then, point the camera optical axis ${\hat{\bm z}}_{\rm cam}[c]$ towards the target (look at position). Afterwards, rotate the transform using the up direction to of the camera, which is $\hat{\bm y}_{\rm cam}[c]$. Note that $\hat{\bm y}_{\rm cam}[c]$ is updated, to match a right-hand coordinate system.
\begin{align}{
    {\bf R}[c]=
    \begin{bmatrix}
        \hat{\bm y}_{\rm cam}[c] \times \frac{-{\bm u}_{\text{cam}}[c]}{|{\bm u}_{\text{cam}}[c]|} ,\;\; \frac{-{\bm u}_{\text{cam}}[c]}{|{\bm u}_{\text{cam}}[c]|} \times \hat{\bm y}_{\rm cam}[c] \times \frac{-{\bm u}_{\text{cam}}[c]}{|{\bm u}_{\text{cam}}[c]|}
        ,\;\; \frac{-{\bm u}_{\text{cam}}[c]}{|{\bm u}_{\text{cam}}[c]|}
    \end{bmatrix}.}
    \label{eq:r_ocs2enu}
\end{align}

\subsection{Pointing noise}
\label{sec:noist_pointing}

\begin{figure}[]
  \centering
  \includegraphics[height=5.8cm]{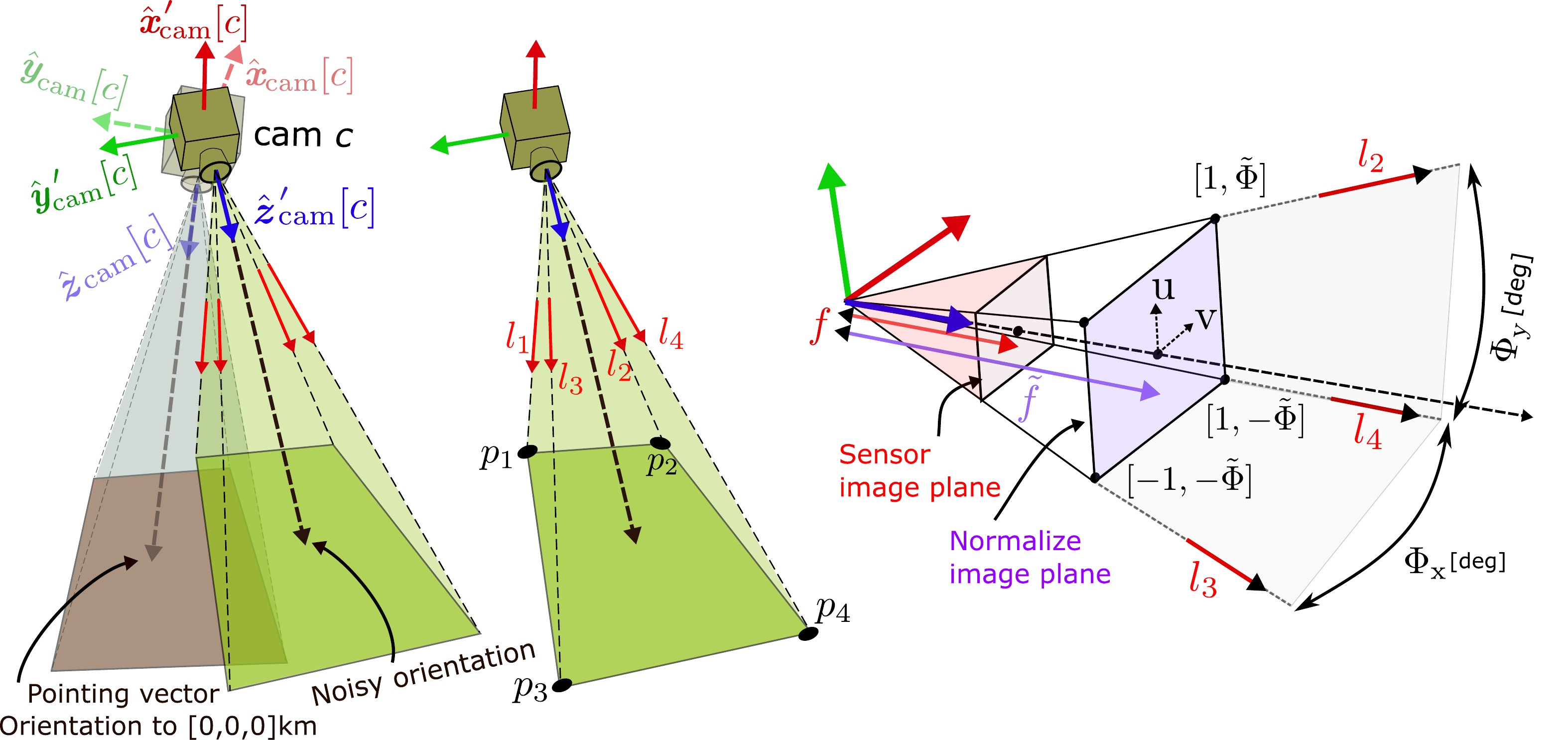}
  \caption{ [Left] The camera coordinate system and the effect of noisy pointing. [Middle] Four rays bound the FOV frustum in 3D. [Right] The camera FOV frustum, image sensor plane, normalized image plane and angular FOV parameters.}
  \label{fig:orient_noise}
\end{figure}

Due to pointing errors of the platform carrying the camera, the orientation of each camera may deviate from that desired. Let $\bar{\bf R}[c]$ be a noisy rotation matrix of camera $c$ relative to the setup coordinate system.
Let $\bar{\bm \upsilon}$ be a noisy version of the vector ${\bm \upsilon}$. Using Eq.~(\ref{eq:omega_exam}), 
\begin{equation}
\begin{split}
    \bar{\bm \upsilon} &= \bar{\bf R}[c]  \;{\bm \upsilon}^{\rm cam}[c].
\end{split}
\label{eq:omega_exam_noise}
\end{equation}
From \cref{eq:omega_exam}, ${\bm \upsilon}^{\rm cam}[c] = {\bf R}[c]^{\top}  \;{\bm \upsilon}$. So,
\begin{equation}
    \bar{\bm \upsilon} = \bar{\bf R}[c]\;{\bf R}[c]^{\top} \; {\bm \upsilon}=
    \delta {\bf R}[c] \; {\bm \upsilon},
\label{eq:qqq2}
\end{equation}
where 
\begin{equation}
    \delta {\bf R}[c] = \bar {\bf R}[c] \; {\bf R}[c]^{\top},\;\;\;\;\bar {\bf R}[c]=\delta {\bf R}[c]\; {\bf R}[c].
\label{eq:R_rel}
\end{equation}
The matrix $\delta {\bf R}[c]$ transforms $\big[ {\hat{\bm x}}_{\rm cam}[c],\,{\hat{\bm y}}_{\rm cam}[c],\,{\hat{\bm z}}_{\rm cam}[c] \big]$ to noisy axes of the camera $\big[ {\hat{\bm x}'}_{\rm cam}[c],\,{\hat{\bm y}'}_{\rm cam}[c],\,{\hat{\bm z}'}_{\rm cam}[c] \big]$ (Fig.~\ref{fig:orient_noise}[Left]). 

To simulate pointing noise, we sample $\delta {\bf R}[c]$ as follows: First, we sample a vector ${\bm \omega}=[\omega_{x},\omega_{y},\omega_{z}]^{\top}$ from a uniform distribution on a unit sphere.\footnote{If the pointing error model has a preferred orientation of ${\bm \omega}$, an anisotropic approach should be used: ${\bm \omega}$ is to be sampled from an anisotropic distribution.} 
Then, we sample an angle $\Theta$ from a Gaussian distribution around ${\bm \omega}$. The standard deviation of the angle $\Theta$ [degrees] is the APE. Therefore, 
  $\Theta  \sim \mathcal{N}\left( 0, {\rm APE}\right)$.

Let $C={\rm cos}\Theta$ and $S={\rm sin}\Theta$.
The rotation matrix $\delta {\bf R}[c]$ is a function of $\Theta$ and ${\bm \omega}$,
and it is calculated by Rodrigues’ formula \cite{blanco2021tutorial,szeliski2022computer},
\begin{align}{
\resizebox{.97\hsize}{!}{$\delta {\bf R}(\Theta,{\bm \omega})[c]=\begin{bmatrix} C + \omega_{x}^2(1-C),\;& \omega_{x}\omega_{y}(1-C)-\omega_z S ,\;& \omega_{x}\omega_{z}(1-C)+\omega_y S\\ 
\omega_{x}\omega_{y}(1-C)+\omega_z S, & C + \omega_{y}^2(1-C), & \omega_{y}\omega_{z}(1-C)-\omega_x S\\ 
\omega_{x}\omega_{z}(1-C)+\omega_y S, & \omega_z \omega_y(1-C)+\omega_x S &, C + \omega_{z}^2(1-C)
\end{bmatrix}.$}}\label{eq:r}
\end{align}
To recap, noisy pointing is simulated as follows. First, we calculate the camera axes vectors in the setup coordinate system $\big[ {\hat{\bm x}}_{\rm cam}[c],\,{\hat{\bm y}}_{\rm cam}[c],\,{\hat{\bm z}}_{\rm cam}[c] \big]$ at ideal pointing, using Eq.~(\ref{eq:r_ocs2enu}). 
Then, Eqs.~(\ref{eq:R_rel},\ref{eq:r}) create random rotations 
$\delta {\bf R}[c]$, $\bar {\bf R}[c]$. This yields perturbed camera axes vectors relative to the setup coordinate system,
\begin{equation}
\label{eq:w_tag_final}
    \bm{\hat{z}'}_{\rm cam}[c] =\bar {\bf R}[c] \; [0,0,1]^{\top}, \;\;\;
    \bm{\hat{y}'}_{\rm cam}[c] =\bar {\bf R}[c] \; [0,1,0]^{\top}, \;\;\;
    \bm{\hat{x}'}_{\rm cam}[c] =\bar {\bf R}[c] \; [1,0,0]^{\top}.
\end{equation}


\section{Field of view}
\label{sec:fov}
Sec.~\ref{sec:overlap_sim} introduced the frustum of a camera FOV. This volume intersects with the domain surface, creating a 2D FOV. This section follows these calculations.\footnote{A commonly done, the camera model here is perspective. Intrinsic calibration can correct for distortions.}
Let ${\bf \Phi} = [{\Phi}_{\rm x}, {\Phi}_{\rm y}]$[deg] denote the angular FOV of a camera in its internal row and column dimensions, and let
 $   [{\Phi}_{\rm y}/{\Phi}_{\rm x}] \equiv \tilde{\Phi} \le 1$.
Define a {\em normalized image plane} (Fig.~\ref{fig:orient_noise}[Right]), corresponding to a unit-less focal length
\begin{equation}
    \tilde{f} = 
    \left[
       {\tan} ({\Phi}_{\rm x}/2)
    \right]^{-1}.
\label{eq:example_w}
\end{equation}
In the normalized image plane, pixel coordinates are ${\rm u}\in[-1,1]$ and ${\rm v}\in[-\tilde{\Phi},\tilde{\Phi}]$ as shown in Fig.~\ref{fig:orient_noise}[Right]. 
The intrinsic matrix~\cite{szeliski2022computer}
of the normalized camera is
\begin{equation}
\label{eq:K}
{\bf K} = \begin{bmatrix}
\tilde{f} & 0 & 0 \\ 
0 & \tilde{f} & 0 \\ 
0 & 0 & 1
\end{bmatrix}.
\end{equation}
A pixel at $[{\rm u}, {\rm v}]$ in normalized camera $c$ that is rotated by noisy rotation $\bar{\bf R}[c]$, relates to a vector ${\bm l}[c]$ in the 3D setup coordinate system is
\begin{equation}
{\bm l}[c] = \bar{\bf R}[c]\;{\bf K}^{-1}\; [{\rm u}, {\rm v}, 1]^{\top}
\label{eq:R_invK_uv}.
\end{equation}

The rays that bound the frustum FOV (Fig.~\ref{fig:orient_noise}[Middle]) have direction vectors \{${\bm l}_1[c],{\bm l}_2[c],{\bm l}_3[c],{\bm l}_4[c]$\} that can be calculated using Eq.~(\ref{eq:R_invK_uv}),
\begin{align}
{{\bm l}}_1[c] &= \bar{\bf R}[c]\;{\bf K}^{-1}\; [-1, \tilde{\Phi}, 1]^{\top}, \;
&{{\bm l}}_2[c] = \bar{\bf R}[c]\;{\bf K}^{-1}\; [1, \tilde{\Phi}, 1]^{\top}, \nonumber \\
{{\bm l}}_3[c] &= \bar{\bf R}[c]\;{\bf K}^{-1}\; [-1, -\tilde{\Phi}, 1]^{\top}, 
&{{\bm l}}_4[c] = \bar{\bf R}[c]\;{\bf K}^{-1}\; [1, -\tilde{\Phi}, 1]^{\top}.
\label{eq:hat_ls}
\end{align}
Along a ray, a point is parameterized by $t$ using,
\begin{align}
{\bm l}_1[c] &= {\bm u}_{\text{cam}}[c] + {{\bm l}}_1[c] t, 
&{\bm l}_2[c] = {\bm u}_{\text{cam}}[c] + {{\bm l}}_2[c] t, \nonumber\\
{\bm l}_3[c] &= {\bm u}_{\text{cam}}[c] + {{\bm l}}_3[c] t, 
&{\bm l}_4[c] = {\bm u}_{\text{cam}}[c] + {{\bm l}}_4[c] t.
\label{eq:ls}
\end{align}
For each ray in~\cref{eq:ls}, we calculate the intersection  with the domain surface.
The result is four points on the domain surface, which defines the surface ${\rm FOV}[c]$. If the surface is flat, then ${\rm FOV}[c]$ is a polygon, whose vectors in \cref{fig:orient_noise}[Middle] are \{$[p_1,\;p_2,\;p_3,\;p_4]$\}. 


\begin{figure}[t]
  \centering
  \includegraphics[height=5.1cm]{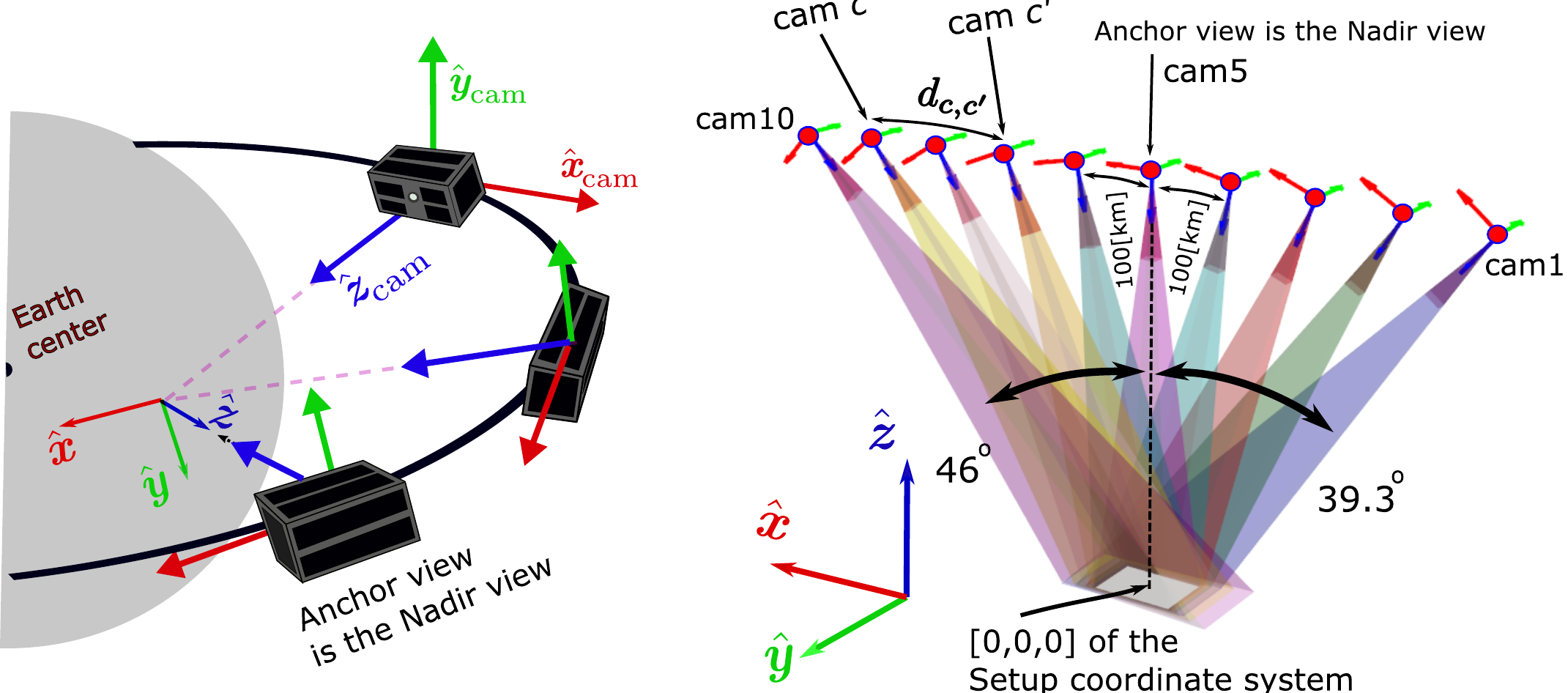}
  \caption{A string of pearls satellite formation setup observing the same domain. [Left] The setup as it is seen from space. [Right] The setup is aligned with the orbital plane.  
  }
  \label{fig:formation_setup}
\end{figure}

\section{Satellites in a formation}
\label{sect:formation}

As a case study, consider simultaneous multi-view imaging of clouds by a formation of small satellites as planned in the CloudCT and C3IEL missions~\cite{cloudct,schilling2019cloudct,tzabari2022advances,dandini20223d}. Multi-view imaging doesn't necessarily need to have a simultaneous setup. Some satellites use sequential acquisition to achieve multi-view. For instance, the Multiangle Imaging SpectroRadiometer~\cite{seiz2006reconstruction} used to measure 3D of clouds, tomography of clouds, and aerosol optical depth. The Mesospheric Airglow/Aerosol Tomography and Spectroscopy~\cite{gumbel2020mats,park2020flight}, the airborne Gimballed Limb Observer for Radiance Imaging of the Atmosphere~\cite{ungermann20113,ungermann2018limited} for tomography of mesoscale gravity waves. The Microwave Limb Sounder~\cite{livesey2006retrieval,wu2006eos} to measures the composition and temperature of the lower atmosphere layers.

The CloudCT setup provides a framework for passive multi-view tomography of the domain. For tomography, multiple satellites should have overlapping FOV near Earths' surface. To enable multiple satellites in an affordable budget, each satellite should be small. Small satellites have limited resources, and usually carry coarser sensors than large satellites. Therefore, nanosatellites have a high APE. 

\begin{figure}[t]
    \centering
    \includegraphics[width=0.99\linewidth]{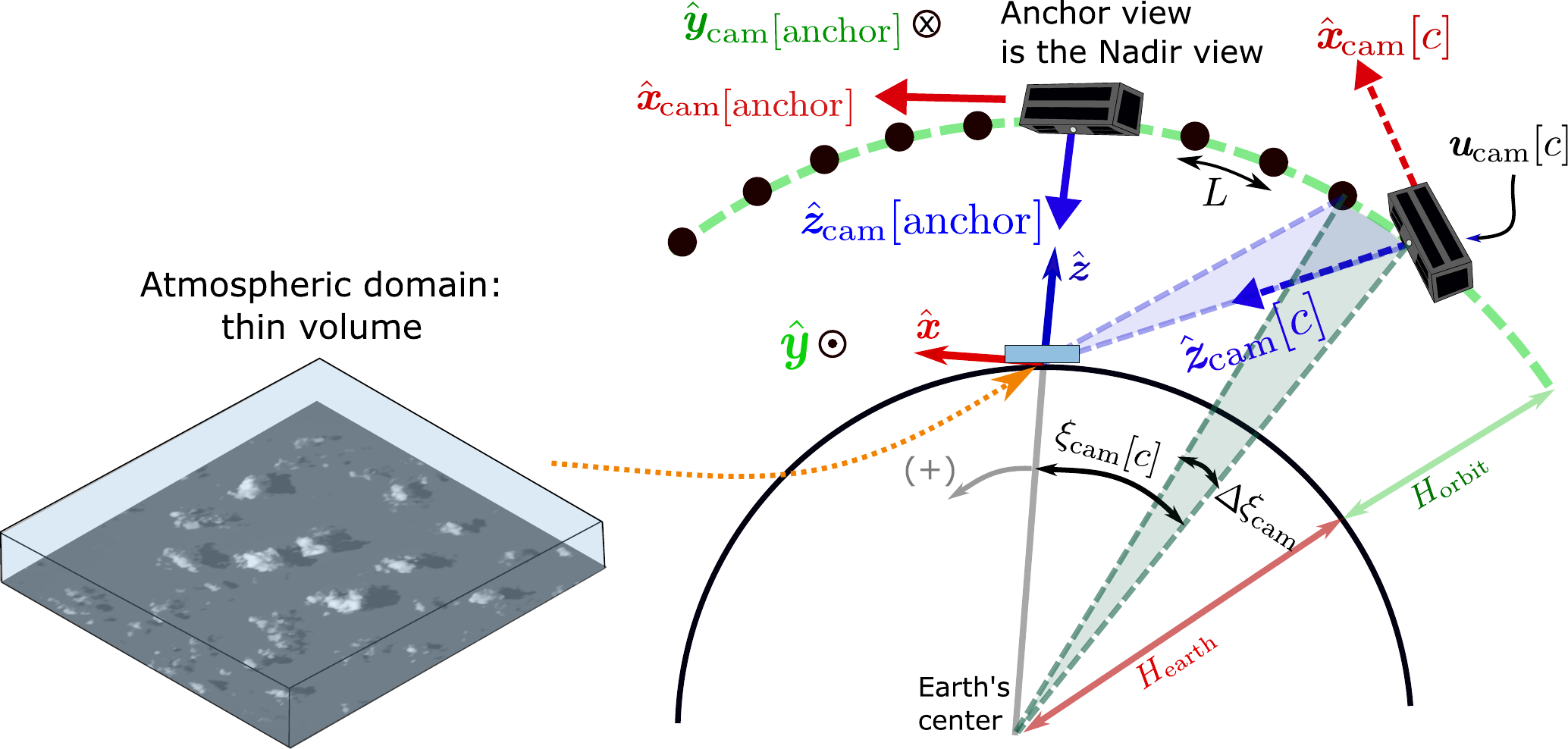}
    \caption{The setup and camera coordinate systems are shown in a simplified (two-dimensional) geometry. The observed domain is thin atmospheric volume. The origin of the domain is on the ground surface.}
    \label{fig:2d-setup}
\end{figure}

Fig. \ref{fig:formation_setup} illustrates a {\em string of pearls} formation: The satellites move consecutively in the same orbit.
Each satellite carries a camera and tries points to the same region as the anchor camera. The anchor, i.e the camera closest to the domain, is nadir-viewing.  The domain zenith is $\bm{\hat{z}}$ (Fig.~\ref{fig:2d-setup}). 
The angle between this zenith and spaceborne camera $c$ is 
$\xi_{\rm cam}[c]$, relative to the center of the Earth. 
Then, $\xi_{\rm cam}[\rm anchor]=0$.

Earth is approximately a sphere having radius $H_{\rm earth}$ $\left[{\rm km}\right]$. The altitude of the orbit is $H_{\rm orbit}$ $\left[{\rm km}\right]$. The orbit radius from the Earth's center is $H=H_{\rm earth}+H_{\rm orbit}$.
The 3D location of a camera in the setup coordinate system is
\begin{align}
    {X}_{\rm cam}[c] = H \,{\rm sin}(\xi_{\rm cam}[c]),\;\;\;
    \;\; {Y}_{\rm cam}[c] = 0, \;\;\;
    {Z}_{\rm cam}[c] = H \,{\rm cos}(\xi_{\rm cam}[c]) - H_{\rm earth}.
\label{eq:coords_sat_TE}
\end{align}
The distance between nearest-neighboring satellites on the orbit arc is denoted by $L$ $\left[{\rm km}\right]$ and it is uniform over the setup. Let $\Delta \xi_{\rm cam} = \frac{L}{H}$.
According to the example in Figs.~\ref{fig:formation_setup},~\ref{fig:2d-setup}, the rightmost satellite has index $c=1$.
For an even $N_{\rm cam}$, 
\begin{equation}
    \xi_{\rm cam}[c] = 
    (c-N_{\rm cam}/2)\Delta \xi_{\rm cam}.
\label{eq:xi_grid}
\end{equation}

Using Eqs.~(\ref{eq:r_ocs2enu},\ref{eq:w_tag_final}), we calculate noisy orientation of each camera $c$ in the setup coordinate system.
As long as the ground FOV is on a scale of $100\times100$ $[{\rm km}]^2$, it is valid to neglect the curvature of the Earth, i.e., assume a regionally flat earth.  The FOV on the ground is then approximately polygonal, as described in \cref{sec:fov}. 
To calculate the overlap terms that intersect the FOV polygons $\Psi$ (\cref{eq:ao}) on the flat ground surface, we used~\cite{shapely2007}.
Since the anchor has nadir-view, its FOV on the domain surface is a rectangle of dimensions ${\rm FOV}[{\rm anchor}]={\rm W}_{\rm x}\times {\rm W}_{\rm y}$, where
\begin{align}
      {\rm W}_{\rm x} = 2 H_{\rm orbit} \; {\rm tan} \Big ( 
\frac{{\Phi}_{\rm x}}{2} \Big ) \; \left[\rm km\right],\;\;\;\;
      {\rm W}_{\rm y} = 2 H_{\rm orbit} \; {\rm tan} \Big ( 
\frac{{\Phi}_{\rm y}}{2} \Big ) \; \left[\rm km\right].
      \label{eq:FOV2FOV}
\end{align}


\cref{sec:overlap_sim} describes criteria for successful geometric self-calibration of $Q\le N_{\rm cam}$ views.
We use the Monte Carlo approach to estimate the probability of meeting these criteria.
A {\em rotation state sample} is a set of random rotations across the formation, $\Omega = \big\{  \bar {\bf R}[c] \big\}_{c=1}^{N_{\rm cam}}$. The state $\Omega$ is sampled by Monte Carlo. The sampled rotation state corresponds to a set of ground FOV polygons $\Psi = \big\{  {\rm FOV}[c] \big\}_{c=1}^{N_{\rm cam}}$. Then, overlap measures are calculated for this sample.  Finally, we use an ensemble of such samples to obtain statistics, i.e., to estimate the probability of successful geometric self-calibration.

\subsubsection{Setup considerations}
\label{sec:consider}

As described in~\cref{sec:overlap_sim}, visual similarity between views $c$ and $c'$ depends on $\mu_{c, c'}$. Here, we chose to work with a distance between the satellites indexed $c$ and $c'$, which we denote $d_{c, c'}$ (Fig.~\ref{fig:formation_setup}). The constraint on $d_{c, c'}$ is similar to the constraint on $\mu_{c, c'}$. If $d_{c, c'}$ is large, the images from $c$ and $c'$ may significantly differ, resulting in very few or null matching feature points between the views.
Let us consider satellites orbiting at $H_{\rm orbit}$=500-600 [km], observing convective cloud fields. Then, as we explain, for sufficient similarity of clouds appearance \cref{sec:overlap_sim} $d_{c, c'}\le 200{\rm [km]}$.

Here is evidence for this limit. Dandini et al. \cite{dandini20223d} showed that at $H_{\rm orbit}=$ 600[km] sufficient cloud matches require $L=150{\rm[km]}$ for which $\mu=$7\textdegree{}.

Zeis et al. \cite{seiz2006reconstruction} use Multiangle Imaging SpectroRadiometer (MISR) data for cloud geometry. The angle between adjacent views in Multiangle Imaging SpectroRadiometer is in the 10\textdegree{}-26\textdegree{} range (relative to the local vertical \cite{johnson2022mapping} as shown in Fig.~\ref{fig:formation_setup}). 

In The CloudCT setup (Fig.~\ref{fig:formation_setup}), $L=100{\rm[km]}$, for a $200{\rm[km]}$ baseline, the range of angles\footnote{The angle between views $c$ and $c'$ is $|{\rm cos}^{-1}(\bm{\hat{z}}_{\rm cam}[c] \cdot \bm{\hat{z}}_{\rm cam}[c'])|$. The angle between views ${\rm cam} 6$ and ${\rm cam} 4$ is $\approx 23$\textdegree{}. The angle between views ${\rm cam} 10$ and ${\rm cam} 8$ is $\approx 14$\textdegree{}.} $\mu=$14\textdegree{}-23\textdegree{}. 

A high angular span of views is essential for a good tomography.
Ref ~\cite{ronen20214d} reports the retrieval quality is poor at total angular spans below 60 [deg].
If the formation orbit is approximately 500 [km], and there are 10 satellites, a high angular span of views, e.g., of 85 [deg], is possible if the distance (on the orbit arc) between nearest-neighbor satellites is around 100 [km].


\begin{figure}[t]
  \centering
  \includegraphics[height=5cm]{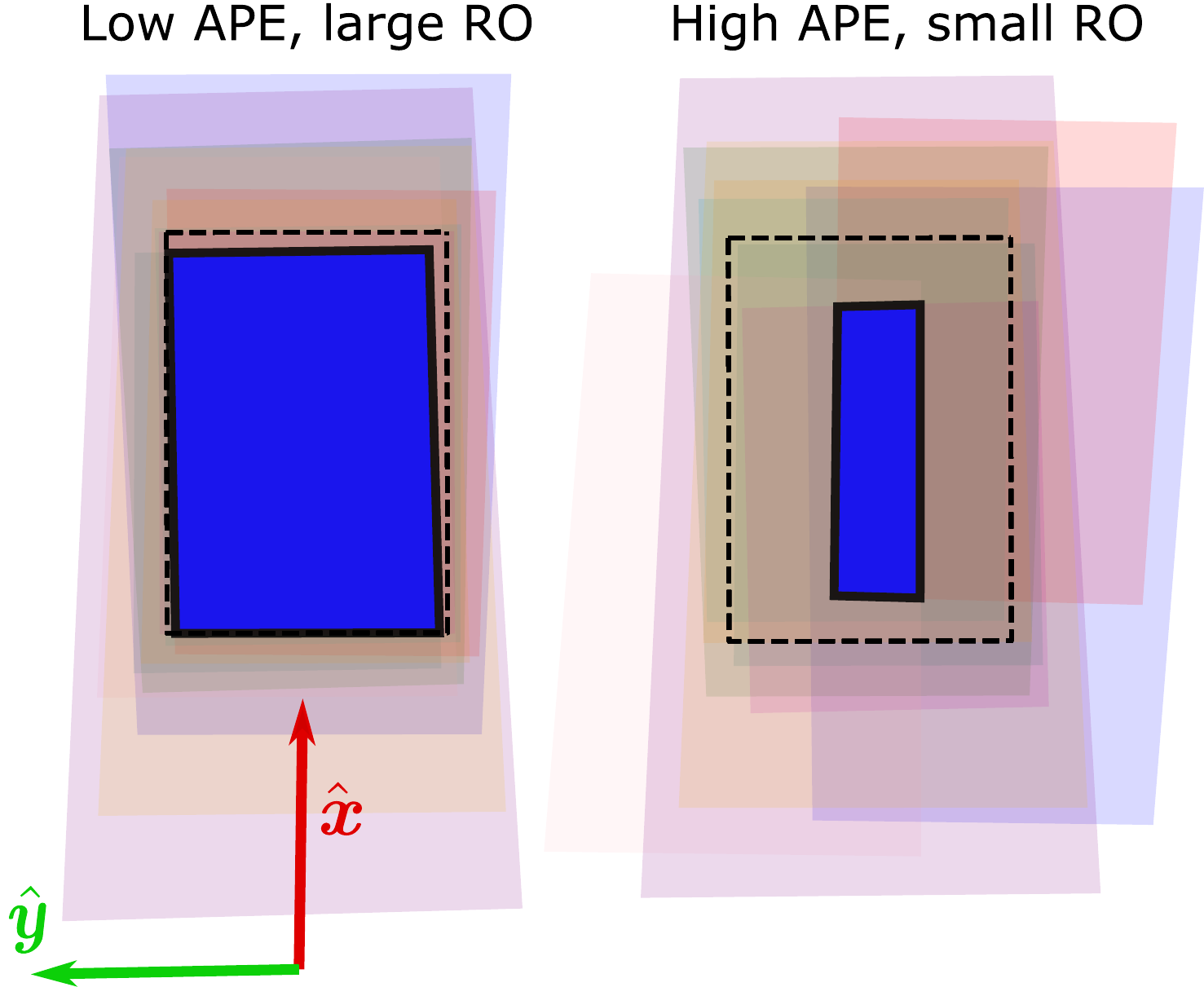}
  \caption{
  Footprints of ten satellites on the ground. Different colors correspond to views by different satellites. The overlap between the views is marked in blue. [Left] Small APE, and large $\overline{\rm RO}$. [Right] High APE, and small $\overline{\rm RO}$. The black dashed rectangle represents the footprint of the anchor view.}
  \label{fig:overlaps_example}
\end{figure}

\begin{figure}[t]
  \centering
  \includegraphics[height=9.5cm]{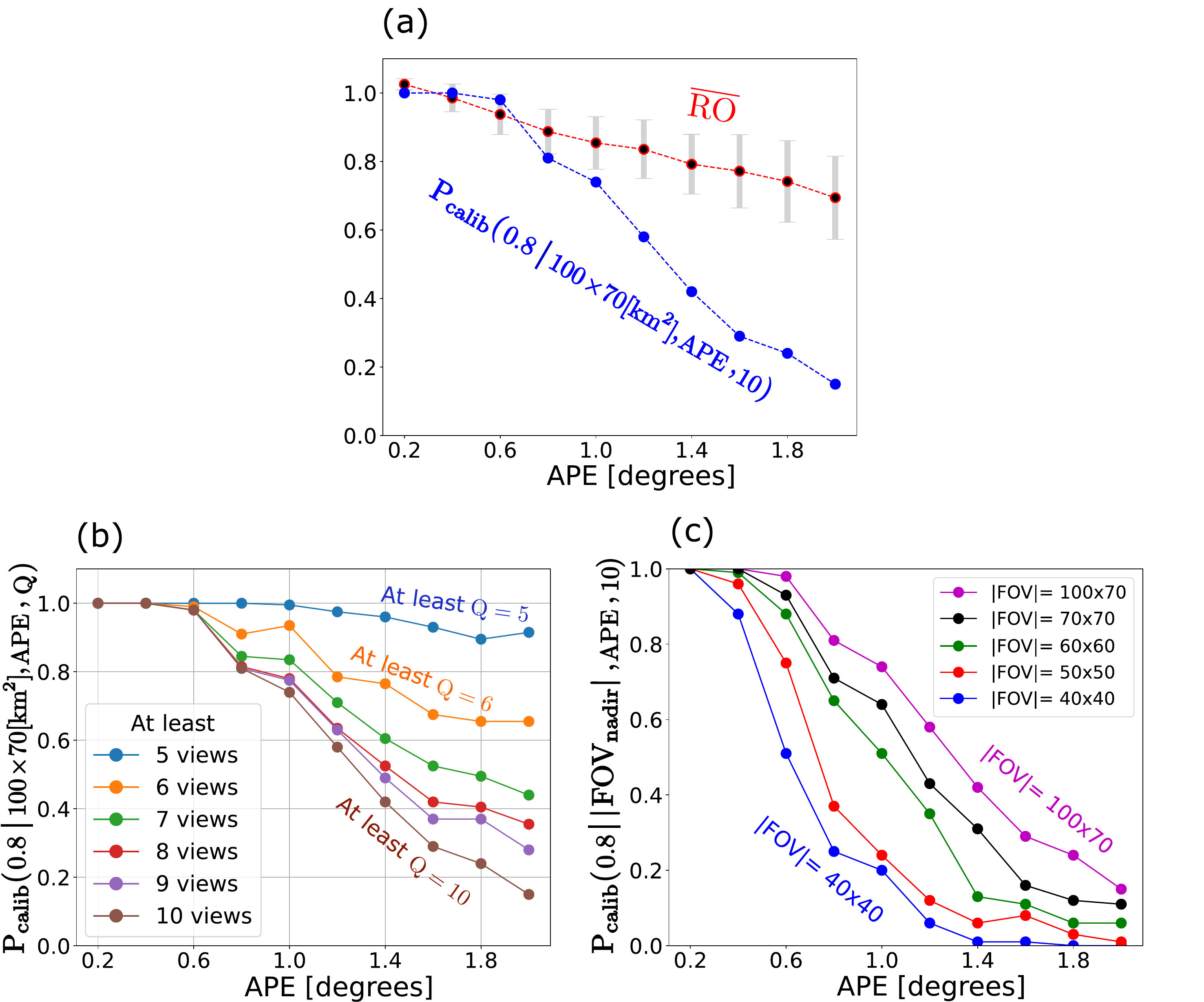}
  \caption{
  [a] The blue plot shows the ${\rm P}_{\rm calib}$ with $T=0.8$ and ${\rm Q}=10$, and the red plot shows $\overline{\rm RO}$. The gray bars represent the standard deviation of the ${\rm RO}$. [b] The ${\rm P}_{\rm calib}$ ($T=0.8$) as a function of $5 \le {\rm Q} \le 10$. Different curves represent different ${\rm Q}$. [a-b] Corresponds to $|{\rm FOV}[{\rm anchor}]|=100 \times 70{\rm km}^2$. [c] The ${\rm P}_{\rm calib}$ ($T=0.8$, ${\rm Q}=10$)
 as a function of the APE and $|{\rm FOV}[{\rm anchor}]|$. Different curves represent different $|{\rm FOV}[{\rm anchor}]|$.}
  \label{fig:FOVS_and_PROBS}
\end{figure}

\section{Results}
\label{sect:resutls}

In this section, we follow the CloudCT mission parameters and analyze the probability of successful geometric self-calibration. Based on \cite{tzabari2022advances}, we set $H_{\rm orbit}=500$ [km], $L$=100 [km], and ${N_{\rm cam}=10}$. Most of the results here assume ${\rm W}_{\rm x}$=100 [km], ${\rm W}_{\rm y}$=70 [km] \cite{von2022cloudct}. 


We set ${\rm T} = 0.8$. For each analyzed case (e.g., different APE, FOV), we sample $N_{\rm MC} = 100$ noisy rotation states. 
\cref{fig:overlaps_example} shows shows how the $\overline{\rm RO}$ changes as a function of APE.
\cref{fig:FOVS_and_PROBS}[a] shows the estimated $\overline{\rm RO}$ for all 10 viewpoints, and ${\rm P}_{\rm calib}$. Both depend on the APE. 
The most significant result is of ${\rm P}_{\rm calib}$; if the APE of satellites is $\approx{2}$\textdegree{} (as typical for nanosatellites) then less than 10\% of all scenes can achieve self-calibration, and consequently tomography, based on all 10 viewpoints. This seems very inefficient. However it is not necessary to use 10 views; also a smaller number of views can enable good science. So what happens if successful analysis can be achieved if {\em at least} 7 out of the 10 viewpoints can achieve self-calibration? for such a question see 
\cref{fig:FOVS_and_PROBS}[b]. 

 It shows that ${\rm P}_{\rm calib}$  increases when ${\rm Q}$ decreases, i.e., it is more likely to perform geometric self-calibration if we tolerate using a subset having fewer overlapping views in the setup. For Q=7, $\approx{50}$\% of scenes will be successfully acquired, self-calibrated and reach tomography, though with varying accuracy.

Suppose we want a better resolution, i.e, a smaller pixel footprint on Earth. Then, for a fixed sensor-array, this can only be achieved using optics that narrows the FOV. Can this be tolerated? Suppose we use a FOV of $40\times{40}[{\rm km}]$, see Fig. \cref{fig:FOVS_and_PROBS}[c]. As seen, if the APE is greater than 1.5\textdegree{}, there is very low chance of success.




\section{Discussion}
\label{sect:conclus}

The paper provides a framework to analyze the probability of successful geometric self-calibration, and consequently multi-view imaging of a domain, 
when cameras pointing is noisy.
The analysis is based on a graph representation.
In the graph, each node is a viewpoint. Two nodes are connected if they satisfy two conditions: (a) a large enough relative overlap between the FOVs they observe in the object domain, through a threshold $T$ and (b) a small difference of feature appearance through a small difference in viewpoint location or direction. This is done by setting a maximum allowable viewing  direction $\mu_{\rm max}$, which is depends of statistics of the scene: in rather flat objects having negligible occlusions, $\mu_{\rm max}$ can be large, but in highly complex 3D structures, such as clouds and trees, $\mu_{\rm max}$ should be small. The paper focused on (a), randomizing the overlap. However, future work can extend this, and randomize scenes whose statistics have varying levels of complexity. 

We apply this analysis to design considerations of a satellite formation.
This implications on the efficiency of the formation, i.e, how many scenes is can successfully capture, relative to the overall number  it {\em tries} to capture. This affects the communication downlink needs. It also affects the achievable resolution: if a wider FOV is needed for a high probability of success, then a fixed sensor pixel array can only support coarser pixel footprint on the ground. 

The paper focused on scenarios where pointing has significant uncertainty, but the location of cameras is well known. This is consistent with the common ability to nail down the location of each camera. This ability exists when cameras are mounted on static positions (building, posts) or when using navigation tools as GPS.  However, the framework can easily extend to general pose errors due to both pointing and location. Moreover, pointing noise need not be isotropic: a perturbation axis may be more or less uncertain than others, and this can be expressed in nonuniform Monte Carlo sampling.  

\section*{Acknowledgments}

We thank Ilham Mammadov and Maximilian von Arnim
for their advices. Yoav Schechner is the Mark and Diane
Seiden Chair in Science at the Technion. He is a Landau
Fellow supported by the Taub Foundation. His work was
conducted in the Ollendorff Minerva Center. Minvera is
funded through the BMBF. This project has received funding
from the European Research Council (ERC) under the
European Union’s Horizon 2020 research and innovation
programme (CloudCT, grant agreement No. 810370), and
the Israel Science Foundation (ISF grant 2514/23).

\bibliographystyle{unsrt}  
\bibliography{references}  


\end{document}